\documentclass[letterpaper, 10 pt, conference]{ieeeconf}  % Comment this line out if you need a4paper
\IEEEoverridecommandlockouts                              % This command is only needed if 
\usepackage{graphics} % for pdf, bitmapped graphics files
\usepackage{epsfig} % for postscript graphics files
\usepackage{times} % assumes new font selection scheme installed
\usepackage{amsmath} % assumes amsmath package installed
\usepackage{amssymb}  % assumes amsmath package installed
\usepackage[hang]{subfigure} % two figures horizontally
\usepackage{cite}
\usepackage{siunitx}
\usepackage{soul}
\usepackage{color}
\usepackage{xcolor}
\usepackage{mathrsfs}
\usepackage{float}
\usepackage{amssymb}
\usepackage{textcomp}
\usepackage{graphicx} % for including images
\usepackage{lipsum}  % for generating filler text

\soulregister\ref7  % 针对\ref命令
\soulregister\cite7 
\soulregister\eqref7
\soulregister\prettyref7

\usepackage{multirow}
\usepackage{algorithm} 
\usepackage{algpseudocode}
\usepackage{makecell}
\usepackage{booktabs}
\usepackage{threeparttable}

\bstctlcite{IEEEexample:BSTcontrol}

\newcommand{\eg}{\textit{e}.\textit{g}.}

\usepackage{hyperref}
\hypersetup{pdfstartview=FitH,
            colorlinks=true,
            linkcolor=red,
            anchorcolor=blue,
            citecolor=black
            }

\title{\LARGE \bf
Prompt, Plan, Perform: LLM-based Humanoid Control \\via Quantized Imitation Learning}
\author{Jingkai Sun$^{1,*}$, Qiang Zhang$^{1,*}$, Yiqun Duan$^{2}$, Xiaoyang Jiang$^{1}$, Chong Cheng$^{1}$ and Renjing Xu$^{1,\dagger}$
\thanks{$^{*}$ are equal contributors, $^{\dagger}$ is the corresponding author}
\thanks{$^{1}$The authors are with The Hong Kong University of Science and Technology (Guangzhou), China.
{\tt\small \{jsun444, qzhang749, ccheng735\}@connect.hkust-gz.edu.cn, renjingxu@hkust-gz.edu.cn} }%
\thanks{$^{2}$The author is with Human $|$ Centric AI Centre, Australia Artificial Intelligence Institute University of Technology Sydney 2007 Ultimo Australia.
{\tt\small yiqun.duan@student.uts.edu.au} }%
}
\begin{document}
\maketitle

\thispagestyle{empty}
\pagestyle{empty}

%%%%%%%%%%%%%%%%%%%%%%%%%%%%%%%%%%%%%%%%%%%%%%[section]%%%%%%%%%%%%%%%%%%%%%%%%%%%%%%%%%%%%%%%%%%%%%%%%%%%%%%%%%%
\begin{abstract}
In recent years, reinforcement learning and imitation learning have shown great potential for controlling humanoid robots' motion. However, these methods typically create simulation environments and rewards for specific tasks, resulting in the requirements of multiple policies and limited capabilities for tackling complex and unknown tasks. To overcome these issues, we present a novel approach that combines adversarial imitation learning with large language models (LLMs). This innovative method enables the agent to learn reusable skills with a single policy and solve zero-shot tasks under the guidance of LLMs.
In particular, we utilize the LLM as a strategic planner for applying previously learned skills to novel tasks through the comprehension of task-specific prompts. This empowers the robot to perform the specified actions in a sequence. To improve our model, we incorporate codebook-based vector quantization, allowing the agent to generate suitable actions in response to unseen textual commands from LLMs. Furthermore, we design general reward functions that consider the distinct motion features of humanoid robots, ensuring the agent imitates the motion data while maintaining goal orientation without additional guiding direction approaches or policies. 
To the best of our knowledge, this is the first framework that controls humanoid robots using a single learning policy network and LLM as a planner. Extensive experiments demonstrate that our method exhibits efficient and adaptive ability in complicated motion tasks.

\end{abstract}

%%%%%%%%%%%%%%%%%%%%%%%%%%%%%%%%%%%%%%%%%%%%%%[section]%%%%%%%%%%%%%%%%%%%%%%%%%%%%%%%%%%%%%%%%%%%%%%%%%%%%%%%%%%

\section{Introduction}
\label{sec: intro}
Advances in humanoid robotics require the development of complex, adaptable, and efficient control policies. These policies enable humanoid robots to operate effectively in human-centered environments, thereby aiding in the successful execution of a diverse range of tasks. Despite significant advancements in the domain of humanoid robotics, a substantial disparity exists between the theoretical potential of these systems and their practical efficacy in executing complex tasks. In this paper, we propose a novel framework to combine the capabilities of Generative Adversarial Imitation Learning (GAIL)\cite{ho2016generative} action data imitation with the planning capabilities of the large language models.

\begin{figure}[t]
    \centering
    \includegraphics[scale=0.42]{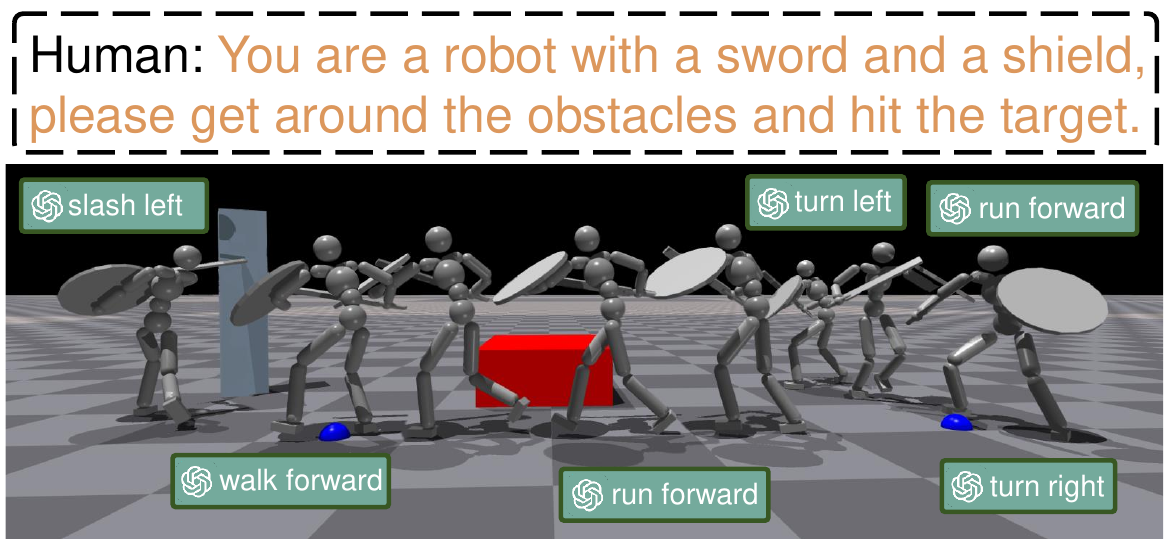}
\caption{Our framework enables robots to combine the skills obtained from imitation learning with the planning capabilities of LLMs to accomplish complex tasks. For example, with known obstacles as well as its own coordinates, the robot accomplishes the task of hitting a target after avoiding obstacles by scheduling reusable skills.}
\label{fig:Fig1}
 \vspace{-5mm}
    
\end{figure}

The GAIL framework consists of two main components: a policy and a discriminator \cite{jolicoeur2018relativistic} and trains policy to perform tasks by imitating expert behavior. The policy interacts with the environment to generate action based on the current state. The discriminator aims to distinguish between the trajectories generated by the expert and those produced by the learning policy. However, merely imitating unlabeled action data limits the learned policy to generating specific actions in particular situations, failing to integrate effectively with high-level policies such as Adversarial Skill Embeddings (ASE)\cite{peng2022ase}. ASE can only accomplish a specific task by training the high-level policy to match the skills of the low-level network in the process of training. 
To address this limitation, we employ motion data in conjunction with its associated textual labels to generate latent vectors\cite{chebotar2021actionable}. These latent vectors are subsequently integrated into the imitation learning network as conditions and skills. These enable the policy to produce contextually appropriate actions when fed with specific skills as input. Leveraging the planning capabilities of LLMs, our method enables the execution of a diverse array of tasks without requiring the development of task-specific high-level networks. This is achieved through the utilization of reusable skills, thereby enhancing the system's adaptability and efficiency.

By utilizing the LLM as a skills planner, our approach enables the humanoid to complete tasks autonomously based on problem descriptions. Even when stringent prompt restrictions are applied to LLMs, its output may not always align with the specified requirements. To guarantee that the robot receives actionable and concise instructions, we refine the complexity of the LLM's outputs by using Contrastive Language-Image Pre-Training (CLIP)\cite{radford2021learning} text encoder and an integrated codebook-based vector quantization. This enhances the robustness of the system concerning the randomness of LLMs-generated outputs. The codebook consists of skills generated from the text labels encoded during training. It is used to correspond similar commands output by the larger model to specific skills. Additionally, we design a general reward consisting of root-oriented and hip-oriented rewards to ensure that the policy trained by imitation learning can execute specific actions along the given directions rather than local coordinates. In this way, we only employ one single policy network to perform specific tasks, without the need for complex hierarchical structures.

Our approach focuses on enhancing the integration of GAIL's dynamic control abilities with the cognitive capabilities of LLMs. In summary, the planning results are shown in Figure~\ref{fig:Fig1} and our contributions include: \textbf{(I)} To the best of our knowledge, we propose a framework that employs a single policy network with LLMs for planning to control humanoid robots to accomplish tasks for the first time. \textbf{(II)} By introducing codebook-based vector quantization, our framework is equipped to handle a variety of similar but unseen instruction sets, enhancing robustness to LLMs or human commands. \textbf{(III)} By designing a general reward for the hip joints, we implemented a single policy to control the direction of robots and perform most of the motion.

%%%%%%%%%%%%%%%%%%%%%%%%%%%%%%%%%%%%%%%%%%%%%%[section]%%%%%%%%%%%%%%%%%%%%%%%%%%%%%%%%%%%%%%%%%%%%%%%%%%%%%%%%%%

\section{Related work}
\label{sec: related}
\subsection{Large Language Model for Robotics}
Recently, the rise of large-scale language models like ChatGPT\cite{ouyang2022training} and Llama2\cite{touvron2023llama} has opened new doors for natural language understanding and generation. Thus, there are various prior works have explored using the large language model for task planning \cite{huang2022inner,huang2022language,di2023towards} or robot manipulation\cite{mees2022calvin,mees2022matters,mees2023grounding,shridhar2022cliport,liang2023code}. SayCan\cite{ahn2022can} combines value functions associated with low-level skills with the large language models, thereby establishing a bridge between rich linguistic understanding and real-world physical interactions. Inner Monologue\cite{huang2022inner} studies the impact of human feedback on LLMs, including scenario descriptions and human-computer interactions. Their findings indicate that detailed examples and closed-loop feedback can improve the performance of LLMs. RT-2\cite{brohan2023rt} employs Internet-scale data to train vision-language models, enabling their application as end-to-end robotic control systems. This model is not merely confined to generating actions based on observations; it also assimilates a comprehensive understanding of the human world derived from online sources. DoReMi\cite{guo2023doremi} facilitates real-time error identification and recovery mechanisms during the execution of plans generated by LLMs on humanoid robots.
\subsection{Adversarial Imitation Learning}
Imitation learning (IL) aims to solve complex tasks where learning a policy from scratch is rather challenging, by learning useful skills or behaviors from expert demonstrations in auto-driving\cite{hu2022model,le2022survey} or robotics\cite{hua2021learning,johns2021coarse,zhu2023viola}. GAIL\cite{ho2016generative} first introduces a discriminator network to provide additional reward signals to the policy. It makes the policy more consistent with the distribution of expert data. Thus, GAIL addresses some of the limitations of behavior cloning \cite{torabi2018behavioral} (such as compounding error caused by covariate shift) by learning an objective function to measure the similarity between the policy and the expert data. Adversarial Motion Prior (AMP)\cite{peng2021amp} learns to imitate different skills from the reference data in order to fulfill high-level tasks. But in AMP, a single policy is typically specialized for a limited set of tasks, because the performance of the agent is strongly correlated with the reward function designed during training. \cite{li2023versatile} realizes a single policy with controllable skill sets from unlabeled datasets containing diverse motion by using a skill discriminator. Based on AMP, ASE\cite{peng2022ase} utilize hierarchical model\cite{pateria2021hierarchical,ji2022hierarchical,huang2022planning,eppe2022intelligent} to make low-level policy learn reusable skills form motion data in latent space. Then it further trains many high-level policies to solve downstream tasks by invoking a low-level policy and designing different reward functions. However, this method still requires many reward functions for specific missions and falls short in its capacity to handle tasks that are not encountered during training. \cite{juravsky2022padl} and \cite{tessler2023calm} employ textual labels and motion data to convert the skills learned through their respective policies into explicit variables. These variables can then be manipulated by humans through syntax trees and finite state machines to facilitate task completion. However, these methods rely on multiple policies and involve considerable manual labor. Based on the above shortcomings, we introduce a framework that incorporates large language models and general directional rewards. This enables the execution of unseen tasks using a single policy network, significantly reducing the manual work required.

%%%%%%%%%%%%%%%%%%%%%%%%%%%%%%%%%%%%%%%%%%%%%%[section]%%%%%%%%%%%%%%%%%%%%%%%%%%%%%%%%%%%%%%%%%%%%%%%%%%%%%%%%%%
 % \vspace{-2.5mm}
\section{Background}
\label{sec:bkg}
In contrast to GAIL, our framework is built upon Conditional Adversarial Latent Models (CALM)\cite{tessler2023calm}, using a conditional discriminator\cite{ma2020ddcgan} to enable it to match a specific latent to an action. Both GAIL and CALM are typically defined in the context of a discrete-time Markov Decision Process (MDP). MDP is defined as a tuple $(\mathcal{S},\mathcal{A},\mathcal{R},p,\gamma)$, where $\mathcal{S}$ is the state space, $\mathcal{A}$ is the action space, $\mathcal{R}$ is the reward function, $p$ represents the probabilities associated with the transition from the current state to the next state for each states-action pair, $\gamma \in [0,1]$ is the discount factor of reward. At each time step $t$, the agent observes the state $s_t \in \mathcal{S}$ from the environment. The agent executes the action $a_t \in \mathcal{A}$ which is sampled from policy $\pi(a_t|s_t)$. Then the state of the agent transitions from $s_t$ to $s_{t+1}$ by $s_{t+1} \sim p(s_{t+1}|s_t,a_t)$. And the agent obtains a reward value at each time step $r_t=\mathcal{R}(s_t,a_t)$. For maximizing the return reward, the objective is to optimize the parameters of the policy $\theta$:
\vspace{-3mm}
\begin{equation}
    \textnormal{arg}\max_{\theta} \mathbb{E}_{(s_t,a_t) \sim p_\theta(s_t,a_t)} \left[ \sum_{t=0}^{T-1} \gamma^t r_t\right]
\end{equation}
where T denotes the time horizon of MDP. 

In the traditional reinforcement learning\cite{polydoros2017survey,mahmood2018benchmarking,kormushev2013reinforcement}, the reward functions need to be manually designed specifically for different tasks, but this is difficult for learning motor skills from expert demonstrations. Imitation learning emerges as a viable alternative for situations where defining a deterministic reward function is impractical, yet the task can be accomplished by imitating expert demonstrations. GAIL tackles some limitations of normal imitation learning such as behavior cloning by introducing a discriminator $\mathcal{D}(s_t,a_t)$ to measure the similarity between a policy and demonstrations. Subsequently, reinforcement learning methods are employed to update $\pi$ in order to optimize the learning objective which requires the state-action pairs of agents. 
% The discriminator $\mathcal{D}(s_t,a_t)$ is trained as a binary classifier to predict whether a given state-action pair $(s_t,a_t)$ originates from samples of the demonstration $\mathcal{M}$ or from the execution of policy $\pi$ by the following objective,
% \begin{equation}
% \begin{aligned}
%     \textnormal{arg}\min_{\mathcal{D}} &-\mathbb{E}_{d^\mathcal{M}(s_t,a_t)}\left[\textnormal{log}(\mathcal{D}(s_t,a_t))\right] \\ & - \mathbb{E}_{d^\pi(s_t,a_t)}\left[\textnormal{log}(1-\mathcal{D}(s_t,a_t))\right]
% \end{aligned}
% \end{equation}
To extend the learning objective to datasets that only offer the state of the agent, we follow \cite{torabi2018generative} to transfer it to the state and the next state pair instead of the original state-action pair. We introduce the Jensen-Shannon divergence\cite{menendez1997jensen} as the objective of the discriminator,
\begin{equation}
\begin{aligned}
   \textnormal{arg}\min_{\mathcal{D}} & -\mathbb{E}_{d^\mathcal{M}(s_t,s_{t+1})}\left[\textnormal{log}(\mathcal{D}(s_t,s_{t+1}))\right] \\ &- \mathbb{E}_{d^\pi(s_t,s_{t+1})}\left[\textnormal{log}(1-\mathcal{D}(s_t,s_{t+1}))\right]
\end{aligned}
\end{equation}
where $d^\mathcal{M}(s_t,a_t)$ and $d^\pi(s_t,a_t)$ denote the state-action pair distribution of reference motion dataset and policy.
%%%%%%%%%%%%%%%%%%%%%%%%%%%%%%%%%%%%%%%%%%%%%%[section]%%%%%%%%%%%%%%%%%%%%%%%%%%%%%%%%%%%%%%%%%%%%%%%%%%%%%%%%%%
\begin{figure}[t]
    \centering
    \includegraphics[scale=0.19]{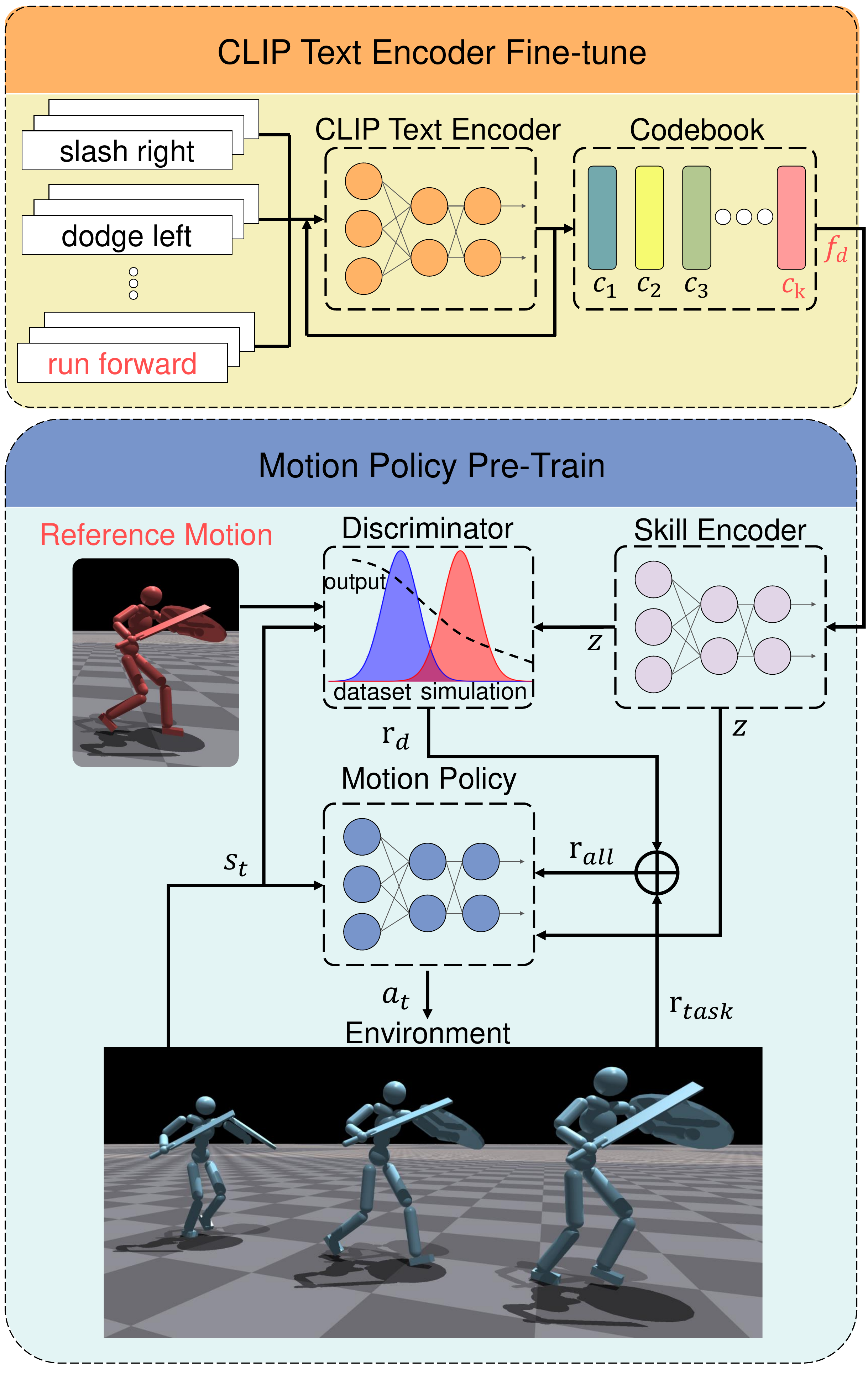}
    \vspace{-3mm}
\caption{\textbf{Overview of our proposed system}. Motion captions with the same semantics are first clustered together by fine-tuning the CLIP Text encoder. Subsequently, the output text features are fed into the policy training by codebook-based vector quantization. Our pre-training system feeds a reference dataset defining the desired underlying motion and its corresponding text labels (marked in red in the figure) into the training discriminator to provide discriminator rewards for policy training. The discriminator reward is then combined with the task reward for controlling orientation to train a policy that allows the robot to execute the demonstrated motion in the specified orientation. These two processes are not trained at the same time.}
\vspace{-5mm}
    \label{Fig2}
\end{figure}

\begin{figure*}[t]
    \centering
    \includegraphics[scale=0.222]{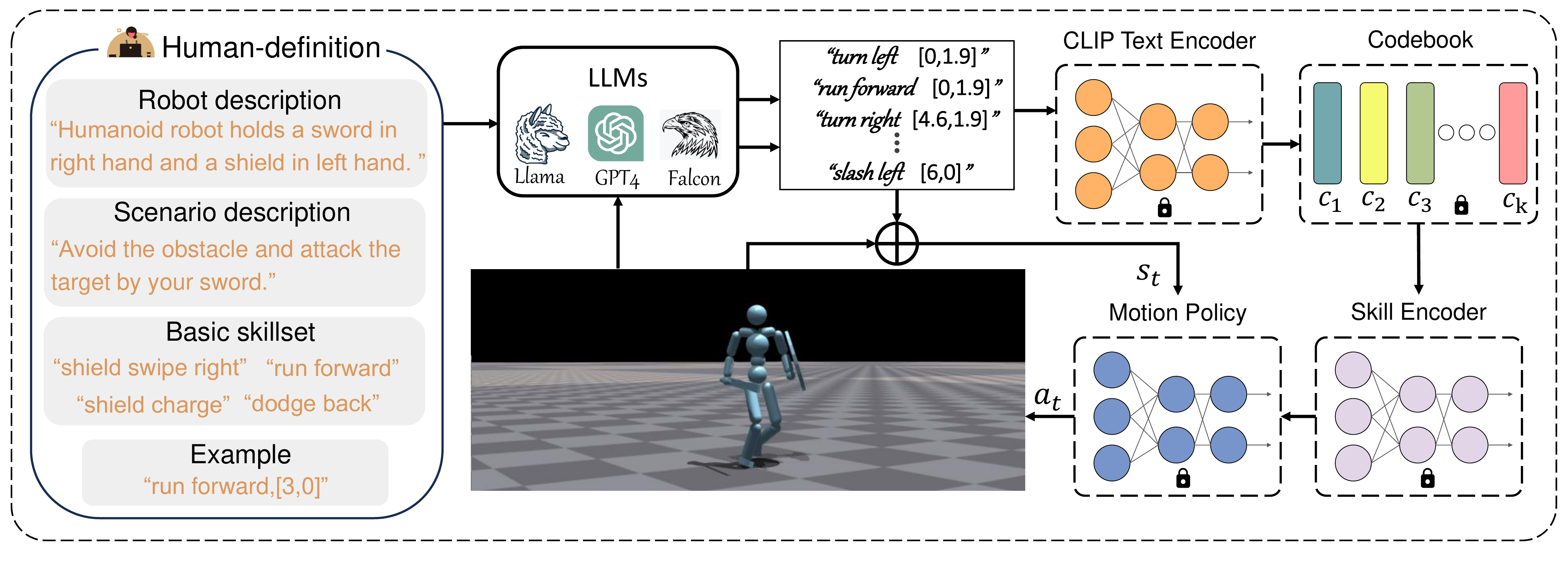}
    \vspace{-5mm}
\caption{\textbf{Evaluation system overview}. The Human definitions are fed into the LLMs as prompts. The LLMs output the sequence of actions and the target orientation of each action. The text of the action sequence is input to the CLIP Text encoder, and the target orientation is input to the policy as an observation concatenated with the observation given by the environment.}
\vspace{-5mm}
    \label{Fig3}
\end{figure*}

\section{APPROACH}

As illustrated in Figure~\ref{Fig2}, the architecture of our proposed method is designed to execute tasks via a single policy framework. Our architecture is composed of three primary components: an adaptive language-based skill motion policy, a CLIP-based adaptive language discrete encoder, and the large language model planner.
\subsection{Adaptive Language-based Skill Motion Policy Pre-train}
\label{sec:raymethod}
Our motion policy pre-train method consists of three main parts: a skill encoder, and a discriminator, a reinforcement learning policy. The skill encoder is used to reduce the dimension of the text features output by CLIP-based adaptive language discrete encoder to generate latent vectors. For generating close and uniform skills of the same kinds of actions on the latent space, we introduce alignment loss and uniformity loss to train the skill encoder like \cite{tessler2023calm}. The discriminator is consistent with the description in the previous section. In our framework, each label of the motion dataset is encoded as a latent vector $z$. We input $z$ to the discriminator and the motion policy as conditions and skills. The discriminator outputs the degree of similarity between the state transitions generated by policy and the dataset. 
% This will form part of the reward of the reinforcement learning framework. The low-level policy will output action based on the state and the latent vector as a command to make a corresponding motion. So we formulate the objective of the discriminator as,
% \begin{equation}
% \begin{aligned}
%    \textnormal{arg}\min_{\mathcal{D}} &-\mathbb{E}_{d^\mathcal{M}(s_t,s_{t+1})}\left[\textnormal{log}(\mathcal{D}(s_t,s_{t+1}|z))\right] \\ &- \mathbb{E}_{d^\pi(s_t,s_{t+1})}\left[\textnormal{log}(1-\mathcal{D}(s_t,s_{t+1}|z))\right]
% \end{aligned}
% \end{equation} 
% Similar to \cite{tessler2023calm}, to overcome the instability in GAIL, conditional discriminator and gradient penalty regularizers\cite{arbel2018gradient} are introduced and the objective is formulated to
\begin{equation}
\begin{aligned}
    \textnormal{arg}\min_{\mathcal{D}} &-\mathbb{E}_{d^\mathcal{M}({s}_t,{s}_{t+1}|)}\left[\textnormal{log}(\mathcal{D}({s}_t,{s}_{t+1}|z))\right] \\ &- \mathbb{E}_{d^\pi(s_t,s_{t+1})}\left[\textnormal{log}(1-\mathcal{D}(s_t,s_{t+1}|z))\right] \\ &+\omega_{gp}\mathbb{E}_{d^\mathcal{M}({s}_t,{s}_{t+1})}\left[\| \mathbf{\triangledown_{\alpha} \mathcal{D}(\alpha)} \|^2\right]
\end{aligned}
\end{equation}
where $\omega_{gp}$ is a hyperparameter to adjust gradient penalty regularizers, $\alpha=(s_t,s_{t+1}|z)$, and $z$ stops the gradient computation in the process to prevent the gradient of the skill encoder from being passed to the learning of the policy. Then the reward of the discriminator is formulated as $r_d=\textnormal{log}(1-\mathcal{D}(s_t,s_{t+1}|z))$. During each iteration, a randomly selected type of motion is input into the discriminator. In this setup, the agent is rewarded more favorably only if its executed actions closely align with the selected motion type.

Unlike CALM, our motion policy is trained further by the combination of two rewards, the discriminator reward and the task reward. This is because the policy trained through imitation learning performs motions in their local coordinates. However, real-world task solutions require the humanoid robot to move in a specified direction, and even upper body action (\eg, sword swinging) needs to determine the orientation. Therefore, the high-level policy is required to specify directions for the low-level policy in the CALM framework. This leads to additional network training and still does not represent the direction of upper body action direction since its network rewards whole body movement direction. Thus, we design a general reward that incorporates the physical mechanism characteristics of humanoid robots. This is aimed at ensuring that the policy trained through imitation learning can move or perform specific actions along specific directions. The general reward consists of three items, root direction (also called pelvis), left hip direction, and right hip direction. The reward is formulated as,
\begin{equation}
    \begin{aligned}
    r_{task}=\omega_{1}\textbf{d}_{root} \cdot \textbf{d}_{t}+\omega_{2}\textbf{d}_{l} \cdot \textbf{d}_{t} + \omega_{2}\textbf{d}_{r} \cdot \textbf{d}_{t}
    \end{aligned}
\end{equation}
where $\omega_{1}$ and $\omega_{2}$ are the weight of the root direction reward and hip direction reward. $\textbf{d}_{root}$, $\textbf{d}_{l}$, and $\textbf{d}_{r}$ are the direction vectors of the root, left hip, and right hip or humanoid robot. 
The total reward is,
\begin{equation}
    r_{all}=\omega_{task}r_{task}+\omega_{dis}\textnormal{log}(1-\mathcal{D}(s_t,s_{t+1}|z))
\end{equation}
where $\omega_{task}$ is the weight of the task reward, and $\omega_{dis}$ is the the weight of discriminator reward.
We find the direction of the root infers the upper body action direction, and the average of two hips direction infers the movements of the whole body. Thus, we design the general reward to ensure that the whole body moves forward (except for movements to the left or right in the local coordinates) and the upper body's action in a specific direction.

\subsection{CLIP-based Adaptive Language Discrete Encoder}
To encode natural language commands(also called captions), our system employs a pre-trained text encoder from the CLIP model. This encoder transforms the input commands into a 512-dimensional vector space. Leveraging an extensive dataset comprising action-related text, we fine-tune this encoder to optimize its performance specifically for our target application. In the fine-tuning phase, we utilize Mean Squared Error (MSE) as a loss function to make the encoded latent vectors obtained from different labels corresponding to the same motion, closer together in the latent space. The loss function is formulated as,
 \vspace{-1mm}
\begin{equation}
\text{MSELoss} = \frac{1}{D} \sum_{i=1}^D (Enc_{T}(text) - Enc_{T}(label))^2
\end{equation}
 \vspace{-1mm}
where $Enc_{T}$ is the text encoder of CLIP, the texts are the input captions of motion, the labels are the text types of these captions, and $D$ is the dimension of text features encoded by the text encoder. For example, if the input caption is "foot strike ahead rapidly", the text type should be "kick". Subsequently, we introduce vector quantization that allows the policy to handle unseen natural language commands and reduce the unnecessary computational resource cost during training. Vector quantization refers to the mapping of each vector component to a finite set of values. In the policy training phase, we employ a codebook-based vector quantization to discretize text features. Specifically, given a feature vector $f$, we use the Euclidean distance as a metric to find its closest corresponding code vector in the codebook. This closest vector serves as the quantized representation for $f$. And the number of motion kinds we need to embed in the training process is the size of the codebook. Finally, we vector quantize $f$ as follows,
\begin{equation}
    \begin{aligned}
    \label{eq:codebook}
        k&=\textnormal{arg}\min_{j}\| f-e_j \|
        \\
        f_d&=\textnormal{Codeboo}k(f) = e_k
        \\
        e_k&= Enc_T(label_k)
    \end{aligned}
\end{equation}

where $f = Enc_{T}(text)$, $e \in R^{K\times D}$, $K$ is the size of the codebook, $D$ is the dimension of the continuous embedding and $e_i \in R^{D}$, $i \in 1,\cdots,K$. We set embedding $e_i$ to the label feature generated by fine-tuning the CLIP text encoder. As such, fine-tuning the CLIP text encoder and vector quantization of the text features allows the agent to handle captions that are unseen and reduces the memory of the training process. 

\subsection{Large Language Model Motion Planner}
In this section, we explore the possibility of extracting action planning knowledge from pre-trained language models without further training, simply by using the prompt. We include descriptions of the robot itself, the task scenario, the action skills it can perform, and some information about the output formatting examples and constraints as prompts. The pipeline of the evaluation system with LLMs is shown in Figure~\ref{Fig3}. Furthermore, the most important part of the prompt is an example of accomplishing a simple task (\eg, moving in a specified direction). Otherwise, LLMs need to interact with the user through subsequent interactions in order to output the action sequences correctly. The output commands of LLMs consist of two parts, the textual commands and coordinate orientation. Textual commands are sequences of reusable action skills that a robot follows to perform an action to accomplish a specific task. The coordinate orientation indicates the direction of the target when the robot performs the action. The former is fed into the policy via a fine-tuned CLIP-based text encoder, while the latter is added to the observation vector to control the robot. In the actual completion of the task, we take the goal, obstacles, and global coordinates of the robot as known information. Under this assumption, actions related to the robot's movement are judged to be completed or not in terms of the distance to the goal position. For actions unrelated to movement, the time required to execute the motion serves as the determining factor for action completion.
\begin{figure}[t]
    \centering
    \includegraphics[scale=0.233]{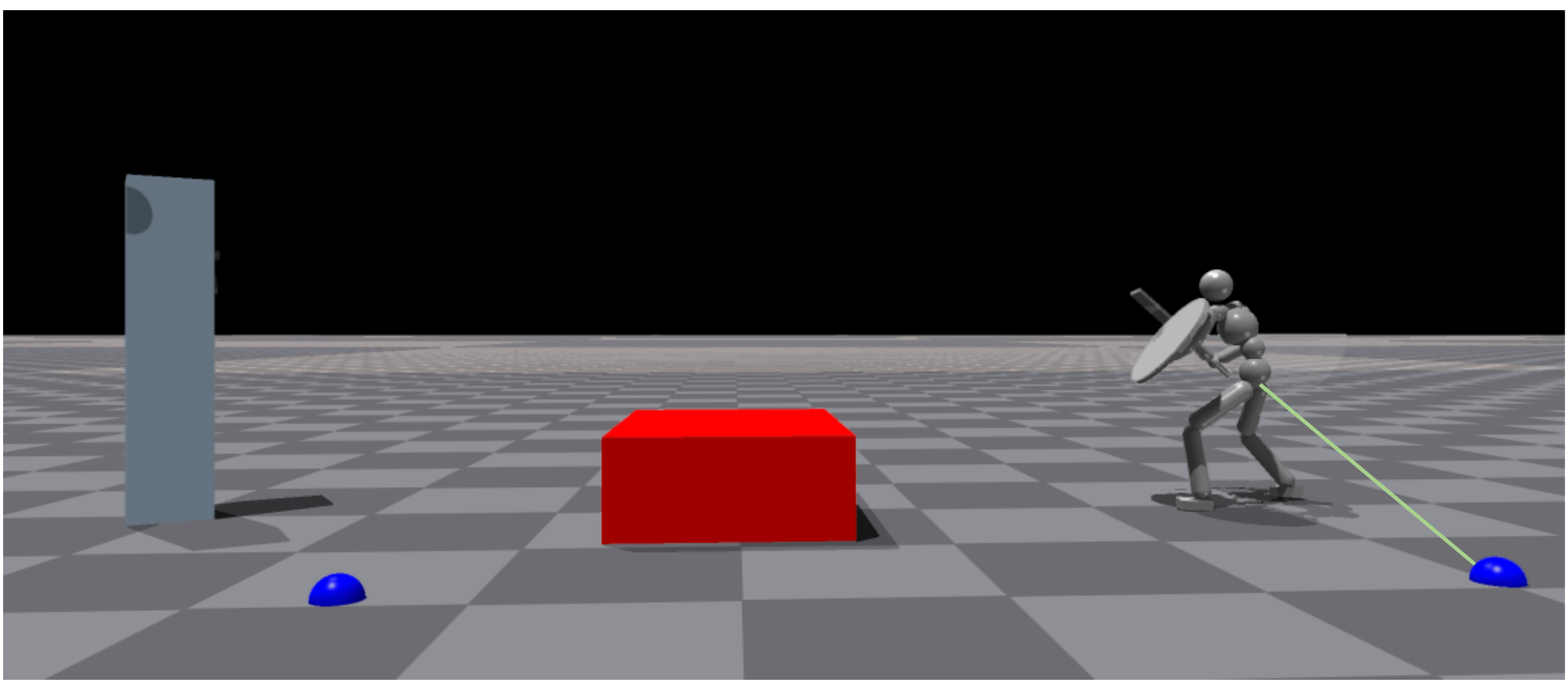}
     \vspace{-1mm}
\caption{\textbf{Initialization of an obstacle avoidance attack task.} The gray rectangle represents the attack target, the blue markers are the middle path point of the LLMs plan, the red is the obstacle, and the green line points to the current target orientation of the robot.}
    \label{Fig4}
\vspace{-4mm}
\end{figure}
 \vspace{-1mm}
\subsection{System Architecture Implementation}
 \vspace{-1mm}
Our robot is a simplified model consisting of spheres, boxes, and cylinders. The whole robot is made up of three joints in the arms, three joints in the legs, a joint in the waist, and a joint in the neck. All of these joints except the knee and elbow have three degrees of freedom. The action space of policy is the target position of each joint. The state space consists of the height of the root from the ground and positions and velocities, rotation, and angular velocities of the other bodies in the robot’s local coordinate. 
% The rotation is composed of the 3D tangent vector and the 3D normal vector for each body part. These vectors offer an alternative way to represent the rotation of each body part, offering numerical stability advantages pertinent to the context in which they are deployed.

%%%%%%%%%%%%%%%%%%%%%%%%%%%%%%%%%%%%%%%%%%%%%%[section]%%%%%%%%%%%%%%%%%%%%%%%%%%%%%%%%%%%%%%%%%%%%%%%%%%%%%%%%%%
\vspace{-2.5mm}
\section{Experiments}
\vspace{-2.5mm}
\label{sec:sdiscussion}
In this section, we conduct experiments involving humanoid robot tasks. These tasks include controlling the robot to perform diverse actions through natural language, incorporating large language models for navigation, and knocking down objects. Specifically, we show that our CLIP-based adaptive language discrete encoder can handle diverse natural language commands. Additionally, the general reward structure we introduce proves effective in directing the robot's movement along a specified direction, without influencing the quality of imitation from the motion dataset. We collect data and train on a single A100 GPU on 4096 Isaac Gym\cite{makoviychuk2021isaac} environments in parallel. And our reinforcement learning algorithm is Proximal Policy Optimization (PPO)\cite{schulman2017proximal}.
\begin{figure*}[t]
    \centering
    \includegraphics[scale=0.29]{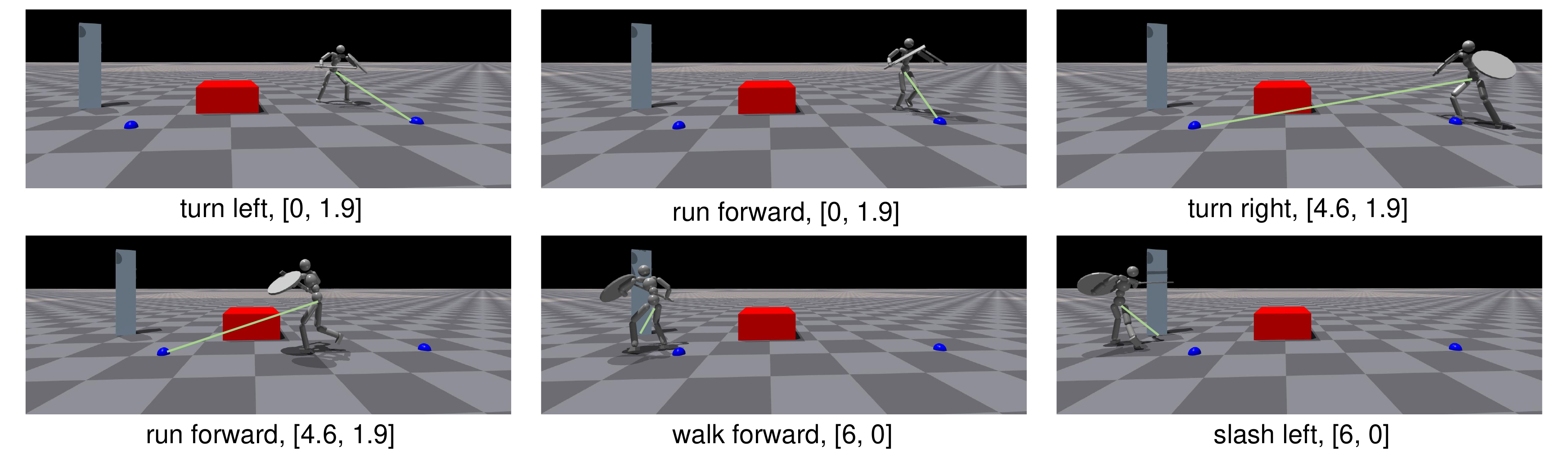}
    \vspace{-3mm}
\caption{\textbf{The movements and their target orientations for each step in completing the obstacle avoidance attack task. }The example shows the initial position of the robot (0,0), obstacle position (3,0), and target position (6,0). The obstacle is a rectangle with a length of 1.2m in the x-axis direction and a length of 1.8m in the y-axis direction. The blue markers are the waypoints of the LLMs plan, the red is the obstacle, and the green line points to the current target orientation of the robot.}
\vspace{-7mm}
    \label{Fig5}
\end{figure*}
\begin{table}[t]
\renewcommand\arraystretch{1.2}
\setlength{\tabcolsep}{5.0pt}
\caption{Comparison of the abilities of different methods}
\begin{center}
\scalebox{1}{
\begin{tabular}{l|cccc}
\hline
\multicolumn{1}{c|}{\multirow{3}{*}{Method}} & \multicolumn{4}{c}{Abilities}                                                                                                                                                                                                                                                                                   \\ \cline{2-5} 
\multicolumn{1}{c|}{}                        & \multicolumn{1}{c|}{\begin{tabular}[c]{@{}c@{}}unseen\\ task\end{tabular}} & \multicolumn{1}{c|}{\begin{tabular}[c]{@{}c@{}}single policy\\ network\end{tabular}} & \multicolumn{1}{c|}{\begin{tabular}[c]{@{}c@{}}input diverse\\ language\end{tabular}} & \begin{tabular}[c]{@{}c@{}}no label\\ request\end{tabular} \\ \hline
AMP\cite{peng2021amp}       & \multicolumn{1}{c|}{}                                                      & \multicolumn{1}{c|}{\checkmark}                                & \multicolumn{1}{c|}{}                                                                 & \checkmark                                  \\
ASE\cite{peng2022ase}       & \multicolumn{1}{c|}{}                                                      & \multicolumn{1}{c|}{}                                                         & \multicolumn{1}{c|}{}                                                                 & \checkmark                                  \\
PADL\cite{juravsky2022padl} & \multicolumn{1}{c|}{\checkmark}                             & \multicolumn{1}{c|}{}                                                         & \multicolumn{1}{c|}{}                                                                 &                                                            \\
CALM\cite{tessler2023calm}  & \multicolumn{1}{c|}{\checkmark}                             & \multicolumn{1}{c|}{}                                                         & \multicolumn{1}{c|}{}                                                                 &                                                            \\
Ours                                         & \multicolumn{1}{c|}{\checkmark}                             & \multicolumn{1}{c|}{\checkmark}                                & \multicolumn{1}{c|}{\checkmark}                                        &                                                            \\ \hline
\end{tabular}
}
\vspace{-8mm}
\end{center}
\label{Table1}
\end{table}
\subsection{Solve Downstream Tasks with LLMs Planner}
In the process of imitation learning, we feed the position, velocity, height of root, rotation relative to the root of each body, the velocity of the joints, and the position of the key bodies (right hand, left hand, right foot, left foot, sword, shield) into discriminator. This input strategy enables the policy to produce actions that utilize fewer dimensions of motion data. In order to demonstrate the motion policy uses reusable skills to accomplish the zero-shot tasks through the LLMs planner, we design a scenario where the robot must navigate around obstacles to reach and knock down a target. As shown in Figure~\ref{Fig4}, the robot is initially oriented squarely towards the target and needs to bypass the red obstacle in the middle and eventually knock down the target once it reaches the vicinity of the target. The LLMs planner, prompted by the pre-input position of the obstacle and other constraints, outputs the robot's actions and corresponding goal orientations. The action type output from the LLMs is fed into the encoder, and the orientation data is incorporated into the observation vector. As the results are shown in Figure~\ref{Fig5}, the complex unseen task can be solved step by step through our proposed method by combining a single reinforcement learning policy with the large language model planner. Table~\ref{Table1} shows the comparison of our approach with other methods. In contrast to ASE, our method does not need to train high-level policies to accomplish specific tasks. Compared to CALM, for scenarios with randomly generated obstacle locations and sizes, our approach does not need to manually design the path points during obstacle avoidance and can automate more complex tasks without the requirement of designing finite state machines.
\begin{figure}[t]
    \includegraphics[scale=0.161]{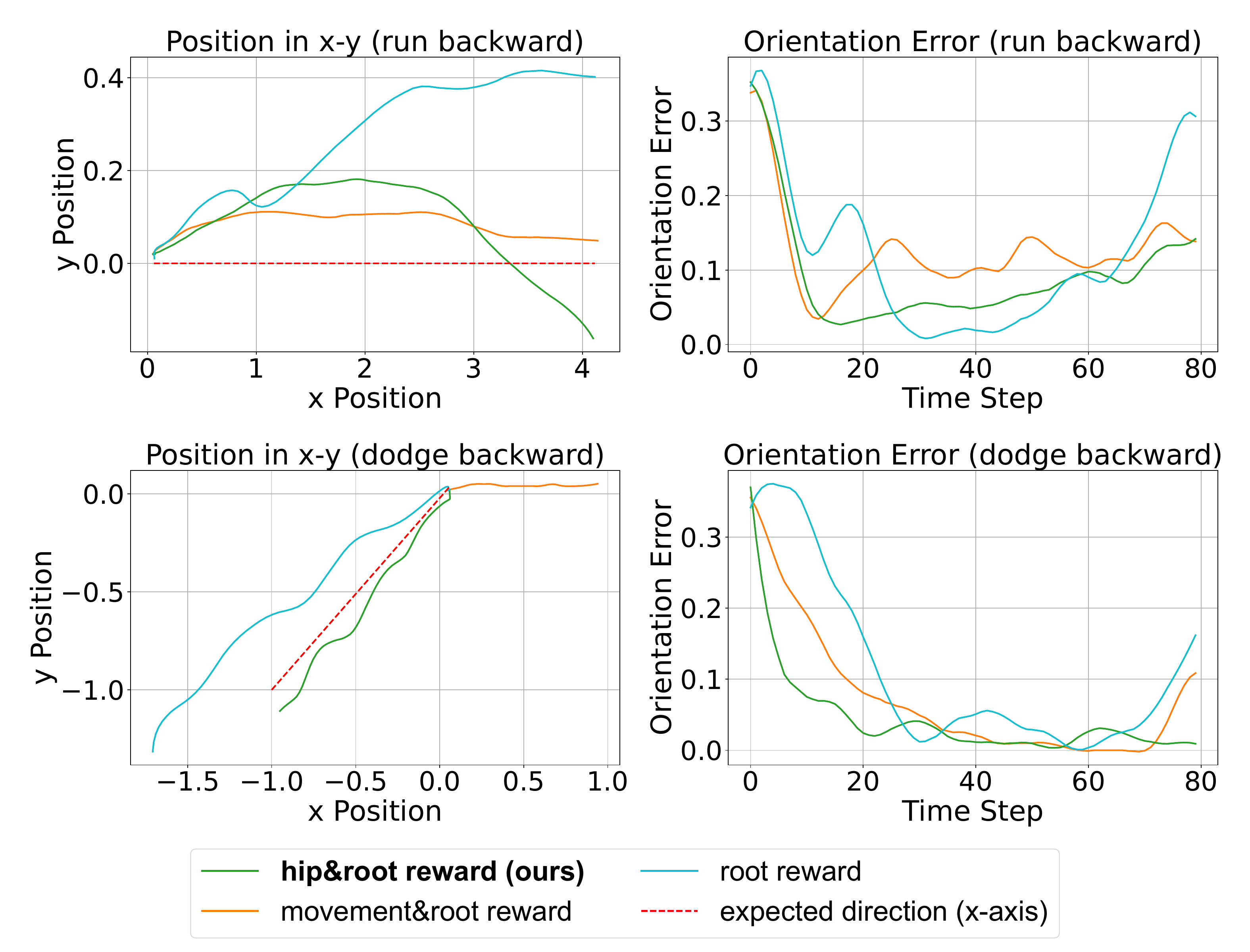}
     \vspace{-5mm}
\caption{Comparisons of Robot Position and Orientation Error (calculated by root direction) during running forward and dodging backward.}
 \vspace{-5mm}
\label{Fig6}
\end{figure}
\begin{table}[t]
\caption{Comparison of the accuracy of different CLIP Text Encoders}

\begin{center}
\scalebox{0.87}{
\begin{tabular}{l|ccllll}
\hline
\multicolumn{1}{c|}{\multirow{2}{*}{Method}} & \multicolumn{6}{c}{Motion class}                                                                                                                                                      \\ \cline{2-7} 
\multicolumn{1}{c|}{}                        & \multicolumn{1}{c|}{slash}   & \multicolumn{1}{c|}{run}     & \multicolumn{1}{c|}{walk}    & \multicolumn{1}{c|}{turn}    & \multicolumn{1}{c|}{dodge}   & \multicolumn{1}{c}{shield} \\ \hline
finetune(our)                                & \multicolumn{1}{c|}{82.45\%} & \multicolumn{1}{c|}{90.21\%} & \multicolumn{1}{l|}{87.93\%} & \multicolumn{1}{l|}{91.93\%} & \multicolumn{1}{l|}{88.13\%} & 79.43\%                    \\
w/o finetune                                 & \multicolumn{1}{c|}{46.23\%} & \multicolumn{1}{c|}{53.78\%} & \multicolumn{1}{l|}{54.34\%} & \multicolumn{1}{l|}{61.93\%} & \multicolumn{1}{l|}{51.20\%} & 45.21\%                    \\ \hline
\end{tabular}
}
\end{center}
\vspace{-7mm}
\label{Table2}
\end{table}
\vspace{-0.5mm}
\subsection{Evaluate Adaptive Language Encoder by Unseen Text}
In the second experiment, we investigate the ability of our method to generalize to unseen commands by using pre-trained language text encoder and codebook-based vector quantization. Specifically, for each representative motion, we test our method with rephrased commands, which are unseen during training. Our test captions consist of three parts: the verbs (\eg, "walk"), adverbs describing the direction (\eg, "forward"), and adverbs describing the features (\eg, "carefully"). Then we generate a large number of related synonyms and combine them randomly as our train dataset and test dataset for fine-tuning the CLIP text encoder. In Table~\ref{Table2}, we show the accuracy of generating appropriate actions in response to unseen commands and also compare the results with the model without fine-tuning. To illustrate, during the policy training phase, a familiar text command such as "run forward" might be used, while an equivalent but previously unseen command could be "rush ahead rapidly". In this experiment, all outputs from the text encoder are fed into codebook-based vector quantization. This step is essential as the policy is designed to manage only those captions that have been encountered during its training phase. To evaluate the system's robustness against novel input, we introduce unseen synonymous captions as test queries. The accuracy rate is calculated based on the network's ability to effectively map previously unseen commands to captions that were encountered during the policy training phase. According to our experiments, our approach can handle unseen commands. This helps to enhance the robustness of our approach, as the output of the large language model is stochastic to some extent, and thus the output of natural language commands needs to be limited to what is acceptable for the model. Moreover, this is also important for future applications of such an approach to direct human use of language control, where user-input commands are much more uncontrollable.
 \vspace{-0.5mm}
\subsection{Comparison of General Task Rewards}
 \vspace{-0.5mm}
In this section, we evaluate the efficacy of three distinct reward structures for controlling the robot's orientation. The first reward structure only focuses on the orientation of the robot's root, the second emphasizes the orientation of the robot's direction of movement, and the third incorporates the orientation of both the robot's root and its two hips. We set the desired direction of the robot along the x-axis. Figure~\ref{Fig6} illustrates the average error of the robot's movement concerning the direction of the target during "dodge backward" and “run forward”. Our experiments suggest that rewarding solely based on root orientation does not offer effective control over the robot's forward movement. Utilizing a reward structure that considers both the robot's direction of movement and root orientation results in a conflict that inhibits the successful execution of the "backward ducking" action. Furthermore, for upper-body motions that do not involve overall robot movement, a movement-based reward causes the robot to persist in moving forward. In contrast, our general reward scheme allows the robot to maintain its initial position during upper-body actions. Therefore, we propose a general reward structure that includes both the root and hip orientations. This structure effectively maintains the desired orientation of the robot's upper body while also facilitating precise control over its direction of movement.
%%%%%%%%%%%%%%%%%%%%%%%%%%%%%%%%%%%%%%%%%%%%%%[section]%%%%%%%%%%%%%%%%%%%%%%%%%%%%%%%%%%%%%%%%%%%%%%%%%%%%%%%%%%
\vspace{-1.5mm}
\section{Conclusion and Limitation}
\vspace{-1.5mm}
\label{sec: conclusion}
In this work, we propose a framework that combines a single adversarial imitation learning policy and LLMs to perform complex tasks by scheduling the skills. Moreover, we design the general reward to ensure that a single control policy is capable of addressing a majority of requirements. Finally, our experiments confirm that the framework we proposed efficiently solves complex tasks and adapts to uncertain semantic outputs of LLMs by introducing codebook-based vector quantization. However, the robot models we used with reference motion data are relatively ideal. Our future work is to generate more practical data and apply the framework to real humanoid robots.
\vspace{-2mm}
\section*{ACKNOWLEDGMENT}
\vspace{-2mm}
We would like to extend our sincere gratitude to Xue Bin (Jason) Peng for his invaluable suggestions and insights that have significantly contributed to the improvement of this project. We also want to extend our heartfelt thanks to the Beijing Innovation Center of Humanoid Robotics Co., Ltd. for their substantial support in our research endeavors. Their assistance has been crucial to the success of our study, providing us with a wealth of resources and valuable collaboration opportunities. We deeply appreciate their guidance and support.
%%%%%%%%%%%%%%%%%%%%%%%%%%%%%%%%%%%%%%%%%%%%%%[section]%%%%%%%%%%%%%%%%%%%%%%%%%%%%%%%%%%%%%%%%%%%%%%%%%%%%%%%%%%

\bibliographystyle{IEEEtran}
\typeout{}
\bibliography{IEEEabrv, mybibfiles}

% Generated by IEEEtran.bst, version: 1.14 (2015/08/26)
\begin{thebibliography}{10}
\providecommand{\url}[1]{#1}
\csname url@samestyle\endcsname
\providecommand{\newblock}{\relax}
\providecommand{\bibinfo}[2]{#2}
\providecommand{\BIBentrySTDinterwordspacing}{\spaceskip=0pt\relax}
\providecommand{\BIBentryALTinterwordstretchfactor}{4}
\providecommand{\BIBentryALTinterwordspacing}{\spaceskip=\fontdimen2\font plus
\BIBentryALTinterwordstretchfactor\fontdimen3\font minus
  \fontdimen4\font\relax}
\providecommand{\BIBforeignlanguage}[2]{{%
\expandafter\ifx\csname l@#1\endcsname\relax
\typeout{** WARNING: IEEEtran.bst: No hyphenation pattern has been}%
\typeout{** loaded for the language `#1'. Using the pattern for}%
\typeout{** the default language instead.}%
\else
\language=\csname l@#1\endcsname
\fi
#2}}
\providecommand{\BIBdecl}{\relax}
\BIBdecl

\bibitem{ho2016generative}
J.~Ho and S.~Ermon, ``Generative adversarial imitation learning,''
  \emph{Advances in neural information processing systems}, vol.~29, 2016.

\bibitem{jolicoeur2018relativistic}
A.~Jolicoeur-Martineau, ``The relativistic discriminator: a key element missing
  from standard gan,'' \emph{arXiv preprint arXiv:1807.00734}, 2018.

\bibitem{peng2022ase}
X.~B. Peng, Y.~Guo, L.~Halper, S.~Levine, and S.~Fidler, ``Ase: Large-scale
  reusable adversarial skill embeddings for physically simulated characters,''
  \emph{ACM Transactions On Graphics (TOG)}, vol.~41, no.~4, pp. 1--17, 2022.

\bibitem{chebotar2021actionable}
Y.~Chebotar, K.~Hausman, Y.~Lu, T.~Xiao, D.~Kalashnikov, J.~Varley, A.~Irpan,
  B.~Eysenbach, R.~Julian, C.~Finn \emph{et~al.}, ``Actionable models:
  Unsupervised offline reinforcement learning of robotic skills,'' \emph{arXiv
  preprint arXiv:2104.07749}, 2021.

\bibitem{radford2021learning}
A.~Radford, J.~W. Kim, C.~Hallacy, A.~Ramesh, G.~Goh, S.~Agarwal, G.~Sastry,
  A.~Askell, P.~Mishkin, J.~Clark \emph{et~al.}, ``Learning transferable visual
  models from natural language supervision,'' in \emph{International conference
  on machine learning}.\hskip 1em plus 0.5em minus 0.4em\relax PMLR, 2021, pp.
  8748--8763.

\bibitem{ouyang2022training}
L.~Ouyang, J.~Wu, X.~Jiang, D.~Almeida, C.~Wainwright, P.~Mishkin, C.~Zhang,
  S.~Agarwal, K.~Slama, A.~Ray \emph{et~al.}, ``Training language models to
  follow instructions with human feedback,'' \emph{Advances in Neural
  Information Processing Systems}, vol.~35, pp. 27\,730--27\,744, 2022.

\bibitem{touvron2023llama}
H.~Touvron, L.~Martin, K.~Stone, P.~Albert, A.~Almahairi, Y.~Babaei,
  N.~Bashlykov, S.~Batra, P.~Bhargava, S.~Bhosale \emph{et~al.}, ``Llama 2:
  Open foundation and fine-tuned chat models,'' \emph{arXiv preprint
  arXiv:2307.09288}, 2023.

\bibitem{huang2022inner}
W.~Huang, F.~Xia, T.~Xiao, H.~Chan, J.~Liang, P.~Florence, A.~Zeng, J.~Tompson,
  I.~Mordatch, Y.~Chebotar \emph{et~al.}, ``Inner monologue: Embodied reasoning
  through planning with language models,'' \emph{arXiv preprint
  arXiv:2207.05608}, 2022.

\bibitem{huang2022language}
W.~Huang, P.~Abbeel, D.~Pathak, and I.~Mordatch, ``Language models as zero-shot
  planners: Extracting actionable knowledge for embodied agents,'' in
  \emph{International Conference on Machine Learning}.\hskip 1em plus 0.5em
  minus 0.4em\relax PMLR, 2022, pp. 9118--9147.

\bibitem{di2023towards}
N.~Di~Palo, A.~Byravan, L.~Hasenclever, M.~Wulfmeier, N.~Heess, and
  M.~Riedmiller, ``Towards a unified agent with foundation models,'' in
  \emph{Workshop on Reincarnating Reinforcement Learning at ICLR 2023}, 2023.

\bibitem{mees2022calvin}
O.~Mees, L.~Hermann, E.~Rosete-Beas, and W.~Burgard, ``Calvin: A benchmark for
  language-conditioned policy learning for long-horizon robot manipulation
  tasks,'' \emph{IEEE Robotics and Automation Letters}, vol.~7, no.~3, pp.
  7327--7334, 2022.

\bibitem{mees2022matters}
O.~Mees, L.~Hermann, and W.~Burgard, ``What matters in language conditioned
  robotic imitation learning over unstructured data,'' \emph{IEEE Robotics and
  Automation Letters}, vol.~7, no.~4, pp. 11\,205--11\,212, 2022.

\bibitem{mees2023grounding}
O.~Mees, J.~Borja-Diaz, and W.~Burgard, ``Grounding language with visual
  affordances over unstructured data,'' in \emph{2023 IEEE International
  Conference on Robotics and Automation (ICRA)}.\hskip 1em plus 0.5em minus
  0.4em\relax IEEE, 2023, pp. 11\,576--11\,582.

\bibitem{shridhar2022cliport}
M.~Shridhar, L.~Manuelli, and D.~Fox, ``Cliport: What and where pathways for
  robotic manipulation,'' in \emph{Conference on Robot Learning}.\hskip 1em
  plus 0.5em minus 0.4em\relax PMLR, 2022, pp. 894--906.

\bibitem{liang2023code}
J.~Liang, W.~Huang, F.~Xia, P.~Xu, K.~Hausman, B.~Ichter, P.~Florence, and
  A.~Zeng, ``Code as policies: Language model programs for embodied control,''
  in \emph{2023 IEEE International Conference on Robotics and Automation
  (ICRA)}.\hskip 1em plus 0.5em minus 0.4em\relax IEEE, 2023, pp. 9493--9500.

\bibitem{ahn2022can}
M.~Ahn, A.~Brohan, N.~Brown, Y.~Chebotar, O.~Cortes, B.~David, C.~Finn, C.~Fu,
  K.~Gopalakrishnan, K.~Hausman \emph{et~al.}, ``Do as i can, not as i say:
  Grounding language in robotic affordances,'' \emph{arXiv preprint
  arXiv:2204.01691}, 2022.

\bibitem{brohan2023rt}
A.~Brohan, N.~Brown, J.~Carbajal, Y.~Chebotar, X.~Chen, K.~Choromanski,
  T.~Ding, D.~Driess, A.~Dubey, C.~Finn \emph{et~al.}, ``Rt-2:
  Vision-language-action models transfer web knowledge to robotic control,''
  \emph{arXiv preprint arXiv:2307.15818}, 2023.

\bibitem{guo2023doremi}
Y.~Guo, Y.-J. Wang, L.~Zha, Z.~Jiang, and J.~Chen, ``Doremi: Grounding language
  model by detecting and recovering from plan-execution misalignment,''
  \emph{arXiv preprint arXiv:2307.00329}, 2023.

\bibitem{hu2022model}
A.~Hu, G.~Corrado, N.~Griffiths, Z.~Murez, C.~Gurau, H.~Yeo, A.~Kendall,
  R.~Cipolla, and J.~Shotton, ``Model-based imitation learning for urban
  driving,'' \emph{Advances in Neural Information Processing Systems}, vol.~35,
  pp. 20\,703--20\,716, 2022.

\bibitem{le2022survey}
L.~Le~Mero, D.~Yi, M.~Dianati, and A.~Mouzakitis, ``A survey on imitation
  learning techniques for end-to-end autonomous vehicles,'' \emph{IEEE
  Transactions on Intelligent Transportation Systems}, vol.~23, no.~9, pp.
  14\,128--14\,147, 2022.

\bibitem{hua2021learning}
J.~Hua, L.~Zeng, G.~Li, and Z.~Ju, ``Learning for a robot: Deep reinforcement
  learning, imitation learning, transfer learning,'' \emph{Sensors}, vol.~21,
  no.~4, p. 1278, 2021.

\bibitem{johns2021coarse}
E.~Johns, ``Coarse-to-fine imitation learning: Robot manipulation from a single
  demonstration,'' in \emph{2021 IEEE international conference on robotics and
  automation (ICRA)}.\hskip 1em plus 0.5em minus 0.4em\relax IEEE, 2021, pp.
  4613--4619.

\bibitem{zhu2023viola}
Y.~Zhu, A.~Joshi, P.~Stone, and Y.~Zhu, ``Viola: Object-centric imitation
  learning for vision-based robot manipulation,'' in \emph{Conference on Robot
  Learning}.\hskip 1em plus 0.5em minus 0.4em\relax PMLR, 2023, pp. 1199--1210.

\bibitem{torabi2018behavioral}
F.~Torabi, G.~Warnell, and P.~Stone, ``Behavioral cloning from observation,''
  \emph{arXiv preprint arXiv:1805.01954}, 2018.

\bibitem{peng2021amp}
X.~B. Peng, Z.~Ma, P.~Abbeel, S.~Levine, and A.~Kanazawa, ``Amp: Adversarial
  motion priors for stylized physics-based character control,'' \emph{ACM
  Transactions on Graphics (ToG)}, vol.~40, no.~4, pp. 1--20, 2021.

\bibitem{li2023versatile}
C.~Li, S.~Blaes, P.~Kolev, M.~Vlastelica, J.~Frey, and G.~Martius, ``Versatile
  skill control via self-supervised adversarial imitation of unlabeled mixed
  motions,'' in \emph{2023 IEEE International Conference on Robotics and
  Automation (ICRA)}.\hskip 1em plus 0.5em minus 0.4em\relax IEEE, 2023, pp.
  2944--2950.

\bibitem{pateria2021hierarchical}
S.~Pateria, B.~Subagdja, A.-h. Tan, and C.~Quek, ``Hierarchical reinforcement
  learning: A comprehensive survey,'' \emph{ACM Computing Surveys (CSUR)},
  vol.~54, no.~5, pp. 1--35, 2021.

\bibitem{ji2022hierarchical}
Y.~Ji, Z.~Li, Y.~Sun, X.~B. Peng, S.~Levine, G.~Berseth, and K.~Sreenath,
  ``Hierarchical reinforcement learning for precise soccer shooting skills
  using a quadrupedal robot,'' in \emph{2022 IEEE/RSJ International Conference
  on Intelligent Robots and Systems (IROS)}.\hskip 1em plus 0.5em minus
  0.4em\relax IEEE, 2022, pp. 1479--1486.

\bibitem{huang2022planning}
S.~Huang, Z.~Wang, J.~Zhou, and J.~Lu, ``Planning irregular object packing via
  hierarchical reinforcement learning,'' \emph{IEEE Robotics and Automation
  Letters}, vol.~8, no.~1, pp. 81--88, 2022.

\bibitem{eppe2022intelligent}
M.~Eppe, C.~Gumbsch, M.~Kerzel, P.~D. Nguyen, M.~V. Butz, and S.~Wermter,
  ``Intelligent problem-solving as integrated hierarchical reinforcement
  learning,'' \emph{Nature Machine Intelligence}, vol.~4, no.~1, pp. 11--20,
  2022.

\bibitem{juravsky2022padl}
J.~Juravsky, Y.~Guo, S.~Fidler, and X.~B. Peng, ``Padl: Language-directed
  physics-based character control,'' in \emph{SIGGRAPH Asia 2022 Conference
  Papers}, 2022, pp. 1--9.

\bibitem{tessler2023calm}
C.~Tessler, Y.~Kasten, Y.~Guo, S.~Mannor, G.~Chechik, and X.~B. Peng, ``Calm:
  Conditional adversarial latent models for directable virtual characters,'' in
  \emph{ACM SIGGRAPH 2023 Conference Proceedings}, 2023, pp. 1--9.

\bibitem{ma2020ddcgan}
J.~Ma, H.~Xu, J.~Jiang, X.~Mei, and X.-P. Zhang, ``Ddcgan: A dual-discriminator
  conditional generative adversarial network for multi-resolution image
  fusion,'' \emph{IEEE Transactions on Image Processing}, vol.~29, pp.
  4980--4995, 2020.

\bibitem{polydoros2017survey}
A.~S. Polydoros and L.~Nalpantidis, ``Survey of model-based reinforcement
  learning: Applications on robotics,'' \emph{Journal of Intelligent \& Robotic
  Systems}, vol.~86, no.~2, pp. 153--173, 2017.

\bibitem{mahmood2018benchmarking}
A.~R. Mahmood, D.~Korenkevych, G.~Vasan, W.~Ma, and J.~Bergstra, ``Benchmarking
  reinforcement learning algorithms on real-world robots,'' in \emph{Conference
  on robot learning}.\hskip 1em plus 0.5em minus 0.4em\relax PMLR, 2018, pp.
  561--591.

\bibitem{kormushev2013reinforcement}
P.~Kormushev, S.~Calinon, and D.~G. Caldwell, ``Reinforcement learning in
  robotics: Applications and real-world challenges,'' \emph{Robotics}, vol.~2,
  no.~3, pp. 122--148, 2013.

\bibitem{torabi2018generative}
F.~Torabi, G.~Warnell, and P.~Stone, ``Generative adversarial imitation from
  observation,'' \emph{arXiv preprint arXiv:1807.06158}, 2018.

\bibitem{menendez1997jensen}
M.~Men{\'e}ndez, J.~Pardo, L.~Pardo, and M.~Pardo, ``The jensen-shannon
  divergence,'' \emph{Journal of the Franklin Institute}, vol. 334, no.~2, pp.
  307--318, 1997.

\bibitem{makoviychuk2021isaac}
V.~Makoviychuk, L.~Wawrzyniak, Y.~Guo, M.~Lu, K.~Storey, M.~Macklin,
  D.~Hoeller, N.~Rudin, A.~Allshire, A.~Handa \emph{et~al.}, ``Isaac gym: High
  performance gpu-based physics simulation for robot learning,'' \emph{arXiv
  preprint arXiv:2108.10470}, 2021.

\bibitem{schulman2017proximal}
J.~Schulman, F.~Wolski, P.~Dhariwal, A.~Radford, and O.~Klimov, ``Proximal
  policy optimization algorithms,'' \emph{arXiv preprint arXiv:1707.06347},
  2017.

\end{thebibliography}
\end{document}